\documentclass[letterpaper]{article} 
\usepackage{aaai2026}  
\usepackage{times}  
\usepackage{helvet}  
\usepackage{courier}  
\usepackage[hyphens]{url}  
\usepackage{graphicx} 
\usepackage{natbib}  
\usepackage{caption} 
\usepackage{amsmath}
\usepackage{amssymb}
\usepackage{booktabs}
\usepackage{multirow}
\usepackage{multicol}
\usepackage{makecell}
\usepackage{tabulary}
\usepackage{float}
\usepackage{algorithm}
\usepackage{algorithmic}
\usepackage{subfiles}
\usepackage{tabularx}
\usepackage{newfloat}
\usepackage{listings}
\usepackage{xcolor}
\usepackage{pifont}       
\usepackage{bbding}       
\usepackage{fontawesome}  
\usepackage{subcaption} 
\DeclareCaptionStyle{ruled}{labelfont=normalfont,labelsep=colon,strut=off} 
\floatstyle{ruled}
\newfloat{listing}{tb}{lst}{}
\floatname{listing}{Listing}
\title{Breaking the Modality Barrier: Generative Modeling for Accurate \\Molecule Retrieval from Mass Spectra}

\author{
    Yiwen Zhang\textsuperscript{\rm 1,2\textdagger},
    Keyan Ding\textsuperscript{\rm 2\textdagger*},
    Yihang Wu\textsuperscript{\rm 1},
    Xiang Zhuang \textsuperscript{\rm 3},
    Yi Yang  \textsuperscript{\rm 2},
    Qiang Zhang\textsuperscript{\rm 4},
    Huajun Chen\textsuperscript{\rm 2,3}\thanks{Corresponding author \\
    {\textdagger}These authors contributed equally to this work.}
}

\affiliations{
    \textsuperscript{\rm 1}The Polytechnic Institute, Zhejiang University \\
    \textsuperscript{\rm 2}ZJU-Hangzhou Global Scientific and Technological Innovation Center, Zhejiang University \\
    \textsuperscript{\rm 3}College of Computer Science and Technology, Zhejiang University \\
    \textsuperscript{\rm 4}ZJU-UIUC Institute, Zhejiang University \\
    \{22460360, dingkeyan, wuyihang, zhuangxiang, y\_yi, qiang.zhang.cs, huajunsir\}@zju.edu.cn
}

\def \OURS{GLMR}
\def \MassRET{MassRET-20k}

\usepackage{bibentry}

\begin{document}

\maketitle

\begin{abstract}
Retrieving molecular structures from tandem mass spectra is a crucial step in rapid compound identification. Existing retrieval methods, such as traditional mass spectral library matching, suffer from limited spectral library coverage, while recent cross-modal representation learning frameworks often encounter modality misalignment, resulting in suboptimal retrieval accuracy and generalization.
To address these limitations, we propose \OURS{}, a Generative Language Model-based Retrieval framework that mitigates the cross-modal misalignment through a two-stage process. In the pre-retrieval stage, a contrastive learning-based model identifies top candidate molecules as contextual priors for the input mass spectrum.
In the generative retrieval stage, these candidate molecules are integrated with the input mass spectrum to guide a generative model in producing refined molecular structures, which are then used to re-rank the candidates based on molecular similarity.
Experiments on both MassSpecGym and the proposed \MassRET{} dataset demonstrate that \OURS{} significantly outperforms existing methods, achieving over 40\% improvement in top-1 accuracy and exhibiting strong generalizability.

\end{abstract}






\section{Introduction}

Tandem mass spectrometry (MS/MS) is one of the most important analytical tools for molecular structure identification~\cite{qiu2023small}. In this technique, target molecules are ionized and undergo a multi-stage fragmentation process, generating a set of fragment-ion signals with specific mass-to-charge ratios (m/z). These signals from the fragment mass spectrum reflect the internal chemical bond cleavage patterns and the functional group distribution characteristics of the molecule. In fields such as metabolomics, natural products discovery, and drug development, the accurate retrieval of molecular structures from MS/MS spectra is a fundamental step toward rapid compound identification~\cite{escher2020tracking}. Herein, the \textit{MS-to-Molecule Retrieval} refers to the process of identifying the most matching molecular structure in a large molecule database based on the input mass spectrum. This retrieval helps researchers quickly locate target compounds~\cite{prudent2021exploring}, eliminating the need for expensive and time-consuming structural analysis experiments~\cite{kind2018identification}.

\begin{figure}[t]
    \centering
    \includegraphics[width=0.5\textwidth,trim=25cm 4cm 32cm 2cm,clip]{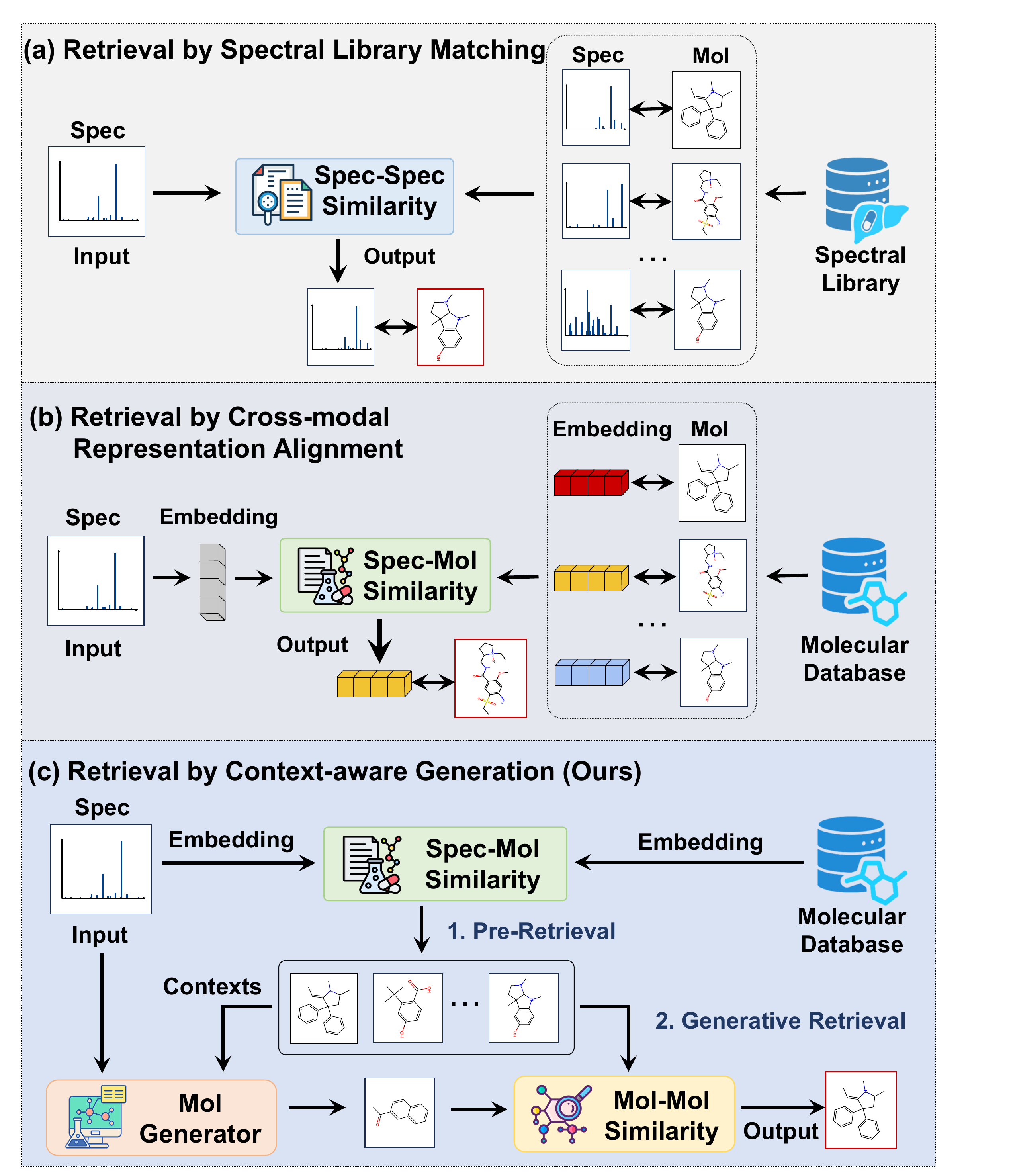}
    \caption{Illustration of the methods for MS-to-Molecule retrieval. (a) \textbf{Spectral library matching method}, where the input mass spectrum is compared against the reference MS of characterized compounds in a database. (b) \textbf{Cross-modal representation alignment method}, where both mass spectra and molecular structures are encoded into a potentially aligned latent space. (c) \textbf{Our method}, which builds upon cross-modal representation alignment, further incorporates a context-aware molecule generator for generative retrieval.
    }
    \label{fig:concept}
\end{figure}

However, inferring molecular structures from MS/MS spectra remains a highly challenging problem~\cite{keifer2017single}. Firstly, different molecules can generate highly similar spectra, while the same molecule may yield substantially different spectral profiles under varying experimental conditions~\cite{el2009mass}. Secondly, real-world spectral data often contain noise, missing peaks, or interfering signals~\cite{wang2005spectral}.
Conventional approaches primarily employ spectral library matching~\cite{stein1994optimization, kwok1973computer, wang2020mass, qin2021deep}, as shown in Figure~\ref{fig:concept}(a), where the experimental spectrum is compared against the reference spectrum of characterized compounds in the databases such as GNPS~\cite{wang2016sharing}, HMDB~\cite{wishart2022hmdb}, MoNA~\cite{vaniya2019massbank} and MassBank~\cite{horai2010massbank}. Although demonstrating reasonable performance for well-documented compounds, these methods exhibit significant limitations due to the restricted coverage of the spectral library~\cite{griss2016spectral}.

Recent advances~\cite{young2024fragnnet, ji2024deepmass, li2024ensemble} have demonstrated the effectiveness of deep learning approaches in directly learning the intricate relationships between spectral patterns and molecular structures. The most prominent approaches~\cite{goldman2023mist, kalia2025jestr} leverage cross-modal representation learning frameworks, as shown in Figure~\ref{fig:concept}(b), where both mass spectra and molecular structures (typically represented as either SMILES strings or molecular graphs) are encoded into a potentially aligned latent space. This paradigm enables efficient MS-to-molecule prediction and retrieval. 
Despite these advancements, a key challenge persists: \textit{modality misalignment}. Mass spectra and molecular structures belong to fundamentally different modalities: the former describes physical fragmentation behavior, while the latter represents chemical structure information. The gap between them makes it difficult to establish a well-aligned representation space. As a result, the current state-of-the-art model JESTR~\cite{kalia2025jestr} demonstrates limited retrieval performance, with top-1 accuracy below 20\% in the MassSpecGym benchmark~\cite{bushuiev2024massspecgym}.

To address this, we propose a generative framework for MS-to-molecule retrieval, termed \textbf{\OURS{}}, as illustrated in Figure~\ref{fig:concept}(c). The core of our approach lies in leveraging a \textit{context-aware generative language model} to bridge the modality gap by generating a molecular structure that is aligned with the input mass spectrum, thereby transforming the cross-modal retrieval into a more tractable unimodal retrieval. Specifically, our method proceeds in two stages:
(1) \textbf{Pre-Retrieval}: A molecule encoder and a spectral encoder are first trained using contrastive learning to retrieve a set of top-ranked candidate molecules, which serve as contextual priors for the input mass spectrum.
(2) \textbf{Generative Retrieval}: These candidates, together with spectral features, guide a generative language model to generate a molecular structure aligned with the input mass spectrum. The generated molecule is then compared with the candidate set via molecular similarity, yielding the final retrieval results. 

To validate the effectiveness of \OURS{}, we conduct evaluations not only on MassSpecGym~\cite{bushuiev2024massspecgym} but also introduce an enhanced MS-to-molecule retrieval dataset, named \textbf{\MassRET{}}, which includes richer spectral variations, providing more challenging and realistic cases. Experimental results on both datasets demonstrate that our method significantly outperforms existing methods.
In summary, the main contributions of this study are as follows: 
\begin{itemize}
    \item We propose a novel MS-to-molecule retrieval framework based on generative language models. Our two-stage approach (pre-retrieval and generative retrieval) effectively alleviates the cross-modal misalignment, improving retrieval accuracy and robustness.
    \item We construct an enhanced molecule retrieval evaluation dataset, enabling comprehensive evaluation of retrieval accuracy, robustness, and generalization.
    \item Our method achieves over 40\% improvement in top-1 accuracy over baseline methods, advancing this field by bridging the gap between mass spectra and molecular structures through generative modeling, enabling more accurate and spectral library-free compound identification.
\end{itemize}

\section{Related Works}

\subsection{MS-to-Molecule Retrieval Methods}
Conventional MS-to-molecule retrieval methods primarily employ spectral library matching, where the input mass spectrum is compared against the reference mass spectrum of known compounds. MASST~\cite{wang2020mass} comprises a web-based system for searching the public data repository within the GNPS/MassIVE knowledge base and an analysis infrastructure for a single mass spectrum.  
DLEAMSE~\cite{qin2021deep} introduces a bioinformatics tool enabling rapid spectral retrieval across public repositories and spectral libraries. However, current approaches are fundamentally constrained by the limited availability of spectrum-molecule pairs, with retrieval performance bounded by spectral library coverage. 
Modeling direct mappings between mass spectra and molecular structures through cross-modal representation learning has emerged as a promising alternative. 
Contrastive learning has become a prevalent strategy for achieving cross-modal alignment~\cite{khosla2020supervised}. For example, MIST~\cite{goldman2023annotating} generates molecular fingerprints based on inferred chemical formulas and performs retrieval via vector similarity, while CMSSP~\cite{chen2024cmssp} integrates molecular graph and fingerprint representations, mapping both spectral and structural modalities into a shared latent space. JESTR~\cite{kalia2025jestr} further enhances contrastive learning with a candidate molecule regularization strategy, and CSU-MS2~\cite{xie2025csu} improves spectral encoding through sinusoidal m/z embeddings and an enhanced attention module, increasing model expressiveness. However, these methods often suffer from modality misalignment, resulting in suboptimal retrieval accuracy.

\subsection{MS-to-Molecule Retrieval Datasets}

Spectral libraries such as GNPS~\cite{wang2016sharing}, MoNA~\cite{vaniya2019massbank}, MassBank~\cite{horai2010massbank}, and NIST~\cite{lemmon2010nist} serve as foundational resources for MS-to-Molecule retrieval by providing experimentally acquired spectra paired with known molecular structures. However, these datasets often suffer from spectral noise, incomplete metadata, or licensing restrictions, which limit their utility for training and evaluating machine learning models. Several standardized benchmarks such as MIST CANOPUS~\cite{goldman2023mist} and CASMI~\cite{schymanski2013casmi} have been proposed, but they are constrained by small size, potential data leakage, or high preprocessing complexity. Recently, MassSpecGym~\cite{bushuiev2024massspecgym} introduced a large-scale, cleaned, and normalized dataset comprising approximately 230k mass spectra, with structurally diverse train-validation-test splits based on MCES~\cite{curchoe2020all} similarity, enabling more robust and reproducible evaluation of retrieval methods.

\section{Methodology}

\begin{figure*}[ht]
    \centering
    \includegraphics[width=1\textwidth,trim=0 15cm 0 9cm,clip]{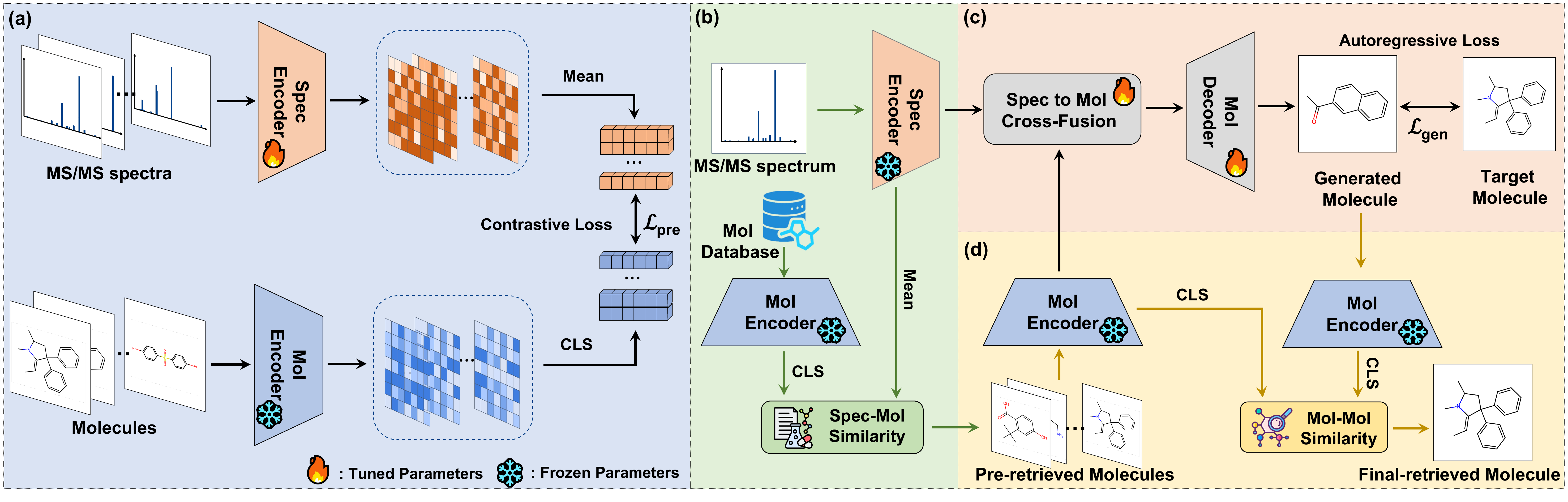}
    \caption{Overview of the proposed \OURS{} method. (a) \textbf{Training process of modality alignment}. We optimize a contrastive loss that encourages mutual alignment between the molecular and spectral modalities. (b) \textbf{Inference process of pre-retrieval}. We use the learned encoders to rank candidate molecules in the retrieval database. The top-$K$ molecules with the highest similarity scores are selected as the output of the pre-retrieval stage. (c)\textbf{ Training process of generative language models}. We leverage a generative language model conditioned on both the input mass spectrum and the prior candidate molecules to produce refined molecular structures. (d) \textbf{Inference process of generative retrieval}. We use the generated molecule to re-rank the pre-retrieved molecules based on molecular similarity.
}
    \label{fig:main}
\end{figure*}

The MS-to-molecule retrieval task aims to rank candidate molecules (from a chemical molecule database) based on a given mass spectrum. 
Formally, given an MS/MS spectrum, the goal is to order a set of candidate molecules such that the correct molecule is positioned at the top~\cite{bushuiev2024massspecgym}.
We address this task through a two-stage retrieval framework (Pre-retrieval and Generative retrieval), as illustrated in Figure~\ref{fig:main}.

\subsection{Pre-retrieval via Cross-modal Representation Alignment}
The first stage performs pre-retrieval by aligning molecular and spectral representations to pick a set of candidate molecules. Specifically, we train a cross-modal alignment model via contrastive learning.
Following the prior work \cite{liu2023multi},  we adopt ChemFormer~\cite{irwin2022chemformer} as the molecular encoder $f_m(\text{Mol};\gamma)$, which is a BART~\cite{lewis2019bart} variant pre-trained on the large-scale ZINC database containing billions of compounds~\cite{irwin2005zinc}. Each input molecule is represented as an SMILES sequence, from which the encoder produces a sequence of token embeddings: 
\begin{align}
    \mathbf{H}^m = f_m(\text{Mol}; \gamma) \in \mathbb{R}^d.
\end{align}
A [CLS] token is prepended to the input sequence, and its final hidden state serves as the global molecular embedding $\mathbf{E}^m = \mathbf{H}^m_{\text{[CLS]}}$.

For the spectral encoder $f_s(\text{spectrum};\eta)$, we employ a Transformer architecture with multi-head attention~\cite{vaswani2017attention}, which allows the model to capture complex relationships across different m/z and intensity dimensions. In contrast to the binning strategies used in previous studies~\cite{kalia2025jestr, chen2024cmssp}, we represent each mass spectrum as a sequence of tuples $(\text{m/z}, \text{intensity})$, where intensity values are normalized to the range $(0,1]$. This sequence is then encoded into a set of hidden representations:
\begin{align}
    \mathbf{H}^s = f_s(\text{Spec}; \eta) \in \mathbb{R}^d.
\end{align}
We apply average pooling over the dimension of sequence length to obtain a fixed-size representation of the mass spectrum, i.e., $\mathbf{E}^s = \frac{1}{T} \mathbf{H}^s$, with $T$ denoting the number of spectral peaks.

The training objective is to align molecular and spectral representations in a latent space. Following the CLIP framework~\cite{radford2021learning}, we optimize a dual-path Info-NCE loss that encourages mutual alignment between the two modalities.
Given a batch of spectrum-molecule pairs $\mathcal{B}$, for the molecule-to-MS alignment, we consider the pairs as the positive sample $\mathbf{E}^s_i$ and construct $N$ negative samples $\mathbf{E}^s_j$ by applying random intensity perturbations to spectral peaks. This yields the molecule-to-MS loss $\mathcal{L}_{mol2ms}$. For the MS-to-molecule alignment, we use the pairs as the positive sample $\mathbf{E}^m_i$ and sample $M$ negative examples $\mathbf{E}^m_j$ from other molecules in the same batch, resulting in the MS-to-molecule loss $\mathcal{L}_{ms2mol}$. The final training loss $\mathcal{L_\text{pre}}$ is computed as the average of these two components, defined as:
\begin{align}
\mathcal{L}_{mol2ms} &= -\frac{1}{\mathcal{\lvert B\rvert}}\sum_{i=1}^\mathcal{\lvert B\rvert} \log \frac{\exp(\frac{\mathbf{E}^m_i\cdot\mathbf{E}^s_i}{\tau})}{\sum_{j=1}^N  \exp(\frac{\mathbf{E}^m_i\cdot\mathbf{E}^s_j}{\tau})}, \\
\mathcal{L}_{ms2mol} &= -\frac{1}{\mathcal{\lvert B\rvert}}\sum_{i=1}^\mathcal{\lvert B\rvert} \log \frac{\exp(\frac{\mathbf{E}^m_i\cdot\mathbf{E}^s_i}{\tau})}{\sum_{j=1}^M \exp( \frac{\mathbf{E}^s_i\cdot\mathbf{E}^m_j}{\tau})}, \\
\mathcal{L_\text{pre}} &= \frac{1}{2}(\mathcal{L}_{ms2mol} + \mathcal{L}_{mol2ms}),
\end{align}
where $\tau$ is the temperature coefficient that controls the sharpness of the similarity distribution.

After training, we use the learned encoders to retrieve candidate molecules in a database based on the cosine similarity between the spectral embedding and each molecular embedding:
\begin{align}
     \mathbf{R}(\mathbf{E}^s,\mathbf{E}^m_i)=\cos(\mathbf{E}^s,\mathbf{E}^m_i)=\frac{\mathbf{E}^s\cdot\mathbf{E}^m_i}{|\mathbf{E}^s|\cdot|\mathbf{E}^m_i|}.
\end{align}
The top-$K$ molecules with the highest similarity scores are selected as the output of the pre-retrieval stage, serving as contextual priors to guide the generation of refined molecules in the next stage.


\subsection{Generative Retrieval via Context-aware Molecule Generation}
The second stage leverages a generative language model conditioned on both the input mass spectrum and the prior candidate molecules to produce refined molecular structures. These generated structures are then used to re-rank the candidate molecules based on molecular similarity, yielding the top-ranked molecules as final retrieval results.

To maintain architectural consistency with the molecular encoder, we employ the ChemFormer Decoder~\cite{irwin2022chemformer} for molecular generation. The input spectrum is encoded as $\mathbf{H}^s$ using the spectral encoder $f_s$, while the top-$K$ candidate molecules from the pre-retrieval stage are encoded as $\mathbf{H}^m_K = \{\mathbf{H}^m_i\}_{i=1}^K$ using the molecular encoder $f_m$, where $K$ denotes the number of pre-retrieved molecules. 
To effectively integrate the spectral and molecular representations, we introduce a \textit{Cross-Fusion} module, which employs cross-attention mechanisms to fuse $\mathbf{H}^s$ and $\mathbf{H}^m_K$. The resulting fused embedding $\mathbf{H}$ is defined as:
\begin{align}
    \mathbf{H}^\text{ca} &= f_{\text{ca}}\left(\mathbf{H}^{s}, \mathbf{H}^m_K; \theta\right) \\
    \nonumber 
    &= \text{Attn}\left(\text{Query}\left(\mathbf{H}^{s}\right), \text{Key}\left(\mathbf{H}^m_K\right)\right) \cdot \text{Value}\left(\mathbf{H}^m_K\right),
\end{align}
where $f_{\text{ca}}$ denotes the cross-attention function parameterized by $\theta$. The Attn function computes cross-attention weights using Query, Key, and Value matrices, which are linear transformations of the input embeddings. This module enables the model to selectively attend to informative molecular candidates while conditioning on the input spectrum. 

During training, both the spectral encoder and molecular encoder are kept frozen to preserve the pre-trained cross-modal alignment learned in the pre-retrieval stage. Training focuses solely on the fusion module and the decoder. The generative model $f_g(\mathbf{H}^\text{ca};\phi)$ autoregressively produces the target molecular structure in the form of a SMILES string $y = (y_1, y_2, \dots, y_Q)$ by maximizing the conditional likelihood of the ground-truth molecule given the fused representation. The training objective is defined as:
\begin{align}
    \mathcal{L_\text{gen}} = -\frac{1}{\mathcal{\lvert B\rvert} \cdot Q}\sum_{i=1}^{\mathcal{\lvert B\rvert}}\sum_{q=1}^{Q} \log P\left(y_{i(q)} \mid y_{i(<q)}, \mathbf{H}^\text{ca}_i\right),
\end{align}
where $P\left(y_{i(q)} \mid y_{i(<q)}, \mathbf{H}^\text{ca}_i\right)$ is the conditional probability of token $y_{i(q)}$ at position $q$, given the preceding tokens $y_{i(<q)}$ and the fused embedding $\mathbf{H}^\text{ca}_i$. 
$Q$ is the target sequence length.

After training, the model is employed to generate a refined molecule, which is then used to re-rank the pre-retrieved molecules based on molecular similarity. Specifically, the generated molecule is encoded as $\mathbf{E}^m_+$, while the candidate molecules are encoded as $\mathbf{E}^m_i$ for $i \in [1, K]$. We calculate cosine similarity between the embeddings of the generated molecule and each candidate:  
\begin{align}\mathbf{R}\left(\mathbf{E}^m_+,\mathbf{E}^m_i\right)=\cos\left(\mathbf{E}^m_+,\mathbf{E}^m_i\right)
     =\frac{\mathbf{E}^m_+\cdot\mathbf{E}^m_i}{|\mathbf{E}^m_+|\cdot|\mathbf{E}^m_i|}.
\end{align}
The candidate molecules are then re-ranked according to their similarity scores, producing the final output of the generative retrieval stage.

\section{Experiments}

\begin{table*}[ht]

\centering
\begin{tabular}{lcccccc}
\toprule
\multicolumn{1}{l}{\multirow{2}{*}{\makecell[l]{Library  Type}}} & \multicolumn{1}{c}{\multirow{2}{*}{Method}} & \multicolumn{3}{c}{Recall$\uparrow$ (\%)} & \multicolumn{1}{c}{\multirow{2}{*}{MRR$\uparrow$ (\%)}} & \multicolumn{1}{c}{\multirow{2}{*}{MCES@1$\downarrow$}}  \\
\cmidrule(lr){3-5} 
 & & Recall@1 & Recall@5 & Recall@20\\
\midrule
\multirow{7}{*}{\makecell[l]{Weight-based  \\ Retrieval Library}} 
 &Random & 0.296 & 1.874 & 7.684  & 1.319 & 31.01\\
 &DeepSets &1.117 &4.049 &13.459 &3.923 &25.47  \\
 &Fingerprint FFN &3.076 &9.211 &22.699 &7.477 &23.85  \\
 &\makecell{DeepSets  + Fourier features} &9.028 &21.081 &38.898 &15.679 &20.87  \\
 & MIST & 18.455 & 40.009 & 64.388  & 29.302 & \underline{15.37}  \\
 & JESTR  & \underline{17.617} & \underline{40.355} & \underline{64.764} & \underline{29.121} & 15.82  \\
 & \OURS{} (Ours) & \textbf{64.172} & \textbf{72.961} & \textbf{78.782}   & \textbf{67.817} & \textbf{11.14}  \\
\midrule
\multirow{7}{*}{\makecell[l]{Formula-based \\ Retrieval Library}} 
 &Random & 2.470 & 10.584 & 21.251 & 5.411 & 13.51 \\
 &DeepSets  &4.699 &12.355 &29.289 &9.901 &13.12 \\
 &Fingerprint FFN &4.978 &15.505 &33.168 &11.193 &13.09  \\
 &\makecell{DeepSets  + Fourier features}  &10.104 &22.015 &40.681 &16.967 &13.01  \\
 & MIST & 10.942 & 23.815 & 44.634  & 18.257 & 12.75  \\
 & JESTR  & \underline{11.772} & \underline{33.258} & \underline{61.006} & \underline{22.825} & \underline{11.73}   \\
 & \OURS{} (Ours) & \textbf{68.478} & \textbf{78.087} & \textbf{84.216}  & \textbf{72.472} & \textbf{5.05}  \\
\bottomrule
\end{tabular}
\caption{Retrieval Performance on \textbf{MassSpecGym}. The best results are in bold, and the results ranked second are underlined.}
\label{tab:retrieval1}
\end{table*}

\begin{table*}[ht]
\centering
\begin{tabular}{lcccccc}
\toprule
\multicolumn{1}{l}{\multirow{2}{*}{\makecell[l]{Library  Type}}} & \multicolumn{1}{c}{\multirow{2}{*}{Method}} & \multicolumn{3}{c}{Recall$\uparrow$ (\%)} & \multicolumn{1}{c}{\multirow{2}{*}{MRR$\uparrow$ (\%)}} & \multicolumn{1}{c}{\multirow{2}{*}{MCES@1$\downarrow$}} \\
\cmidrule(lr){3-5} 
& & Recall@1 & Recall@5 & Recall@20\\
\midrule
\multirow{7}{*}{\makecell[l]{Weight-based  \\ Retrieval Library}} 
 &Random & 0.330 & 1.864  & 7.605 & 1.329  & 30.61 \\
 &DeepSets &0.986 &3.707 &12.540 &3.902 &26.40  \\
 &Fingerprint FFN &3.733 &12.379 &25.705 &8.898 &22.48  \\
 &\makecell{DeepSets + Fourier features} &8.002 &20.467 &38.140 &14.962 &21.69  \\
 & MIST  &14.388 &36.417 &63.526 &25.499 &18.71   \\
 & JESTR  & \underline{16.490} & \underline{38.450} & \underline{60.636} & \underline{27.454} & \underline{18.03}    \\
 & \OURS{} (Ours) & \textbf{54.042} & \textbf{64.347} & \textbf{72.984}  &\textbf{58.835} & \textbf{12.08}   \\
\midrule
\multirow{7}{*}{\makecell[l]{Formula-based \\ Retrieval Library}} 
&Random & 2.012 & 9.094 & 26.139 & 5.956 & 13.51 \\
 &DeepSets  &2.729 &9.116 &27.564 &6.737 &13.39  \\
 &Fingerprint FFN &2.912 &10.573 &28.322 &8.201 &13.15  \\
 &\makecell{DeepSets + Fourier features}  &4.621 &14.063 &34.497 &10.927 &13.11  \\
 & MIST &6.779 &16.882 &34.290 &12.780 &13.04   \\
 & JESTR  & \underline{7.440} & \underline{23.314} & \underline{47.356} & \underline{16.282} & \underline{11.86}   \\
 & \OURS{} (Ours) & \textbf{51.141} & \textbf{60.062} & \textbf{70.671}  & \textbf{55.565} & \textbf{6.94}  \\
\bottomrule
\end{tabular}
\caption{Retrieval Performance on {\MassRET{}}.  The best results are in bold, and the results ranked second are underlined.}
\label{tab:retrieval2}
\end{table*}

This section presents a comprehensive evaluation of the retrieval performance of \OURS{}. We first detail the experimental settings, including datasets, baselines, training configuration, and evaluation criteria. We then report performance on the MassSpecGym and \MassRET{} benchmarks, followed by analyses of modality alignment and generation ability. Lastly, we conduct an ablation study to quantify the contribution of each stage and assess the sensitivity to the number of pre-retrieved candidates.

\subsection{Experimental Settings}

\subsubsection{Datasets}
We evaluate \OURS{} on two benchmark datasets: {MassSpecGym} ~\cite{bushuiev2024massspecgym} and {\MassRET{}}. (1) \textbf{MassSpecGym} provides two retrieval libraries for each MS/MS spectrum. The first is based on molecular \textit{weight} inferred from the precursor m/z, and the second leverages chemical \textit{formula} matching. 
(2) To better evaluate model performance under diverse experimental conditions, we construct a new benchmark dataset, \textbf{\MassRET{}}.  To avoid data leakage, we exclude molecules that appear in the MassSpecGym training set, resulting in a clean evaluation set of approximately 20k spectrum-molecule pairs.  Compared to the metadata of MassSpecGym, which includes only two ionization adducts, and where only 53\% of the data provides normalized collision energy, resulting in incomplete data, our constructed dataset includes 12 ionization adducts, where all entries include normalized collision energy. As a result, our dataset provides more comprehensive information and more accurately reflects real-world scenarios. Furthermore, the same molecule exhibits different mass spectra under different ionization adducts, making \MassRET{} more challenging than MassSpecGym. 
More details of \MassRET{} are described in Appendix \ref{ap:massret}.

\subsubsection{Training Setup}
Our model is trained in two stages using spectrum-molecule pairs from the MassSpecGym training set. In the pre-retrieval stage, we initialize the molecular encoder with pre-trained ChemFormer weights and randomly initialize the spectral encoder. The model is trained for 300 epochs using contrastive loss, with only the spectral encoder updated while the molecular encoder is frozen.
In the generative retrieval stage, we further initialize the decoder with pre-trained weights from ChemFormer and randomly initialize the cross-fusion module. Training runs for 30 epochs with the encoders frozen, updating only the fusion module and decoder to generate molecules conditioned on spectral and contextual information. More training details are provided in Appendix \ref{sec:Implementation_Details}.

\subsubsection{Baselines}
We compare \OURS{} against a range of baselines spanning traditional and deep learning approaches for MS-to-molecule retrieval. These include Fingerprint FFN~\cite{rumelhart1986parallel}, DeepSets~\cite{zaheer2017deep}, DeepSets with Fourier features~\cite{zaheer2017deep}, MIST~\cite{goldman2023mist}, and JESTR~\cite{kalia2025jestr}, the current state-of-the-art method for mass spectrum-based cross-modal molecular retrieval. Their methodological details are provided in the Appendix~\ref{sec:Baselines}.  

\subsubsection{Evaluation Metrics}   
We employ three standard metrics to evaluate MS-to-molecule retrieval performance:
(i) \textbf{Recall@K} measures the proportion of test samples for which the ground-truth molecule appears within the top-$K$ ranked candidates. We report Recall@1, Recall@5, and Recall@20 as percentages.
(ii) \textbf{MRR} (Mean Reciprocal Rank) captures the average inverse rank of the first correct match, giving higher weight to models that rank the true molecule more highly. 
(iii) \textbf{MCES@1} (Maximum Common Edge Subgraph similarity at rank 1) evaluates structural similarity between the top-1 predicted molecule and the ground truth. 
More evaluation metric details are provided in Appendix~\ref{sec:Metrics}. 

\subsection{Main Results}

\subsubsection{Performance on MassSpecGym}
Table~\ref{tab:retrieval1} presents the retrieval performance on the MassSpecGym benchmark. \OURS{} consistently outperforms all baseline methods across all metrics.
On the weight-based and formula-based retrieval tasks, \OURS{} achieves a remarkable improvement in Recall@1, 46\% and 56\% respectively, over the previous state-of-the-art method JESTR. The significant reduction in MCES@1 further demonstrates that the top-1 predictions from \OURS{} are structurally closer to the ground truth, even when the exact match is not retrieved.
This performance leap stems from \OURS{}'s two-stage design, which decouples retrieval into pre-retrieval (cross-modal alignment) and generative retrieval (context-aware generation). While prior methods encode molecules and mass spectra into a potentially aligned latent space (similar to the pre-retrieval in \OURS{}), their cross-modal alignment capability remains limited. \OURS{} addresses this by generating a molecule conditioned on the input mass spectrum and pre-retrieved candidates, thereby reframing the task as a unimodal retrieval process that effectively bridges the modality gap.

\subsubsection{Performance on \MassRET{}}

Table~\ref{tab:retrieval2} shows the generalization ability of all models on the proposed \MassRET{} benchmark, a more challenging and realistic dataset with diverse ionization adducts and complete experimental metadata. All models are trained solely on the MassSpecGym training set and evaluated in a zero-shot~\cite{pourpanah2022review} transfer setting, making this a rigorous test of \textit{generalization}.
As shown in Table~\ref{tab:retrieval2}, \OURS{} remains the top-performing method, significantly outperforming all baselines.
Compared to baselines, \OURS{} shows superior generalization on both retrieval libraries, with notably higher Recall@K and MRR, as well as lower MCES@1 scores. These gains underscore \OURS{}'s ability to generalize to unseen mass spectra and varying ionization conditions.

\begin{figure}[ht]
    \centering
    \includegraphics[width=0.47\textwidth]{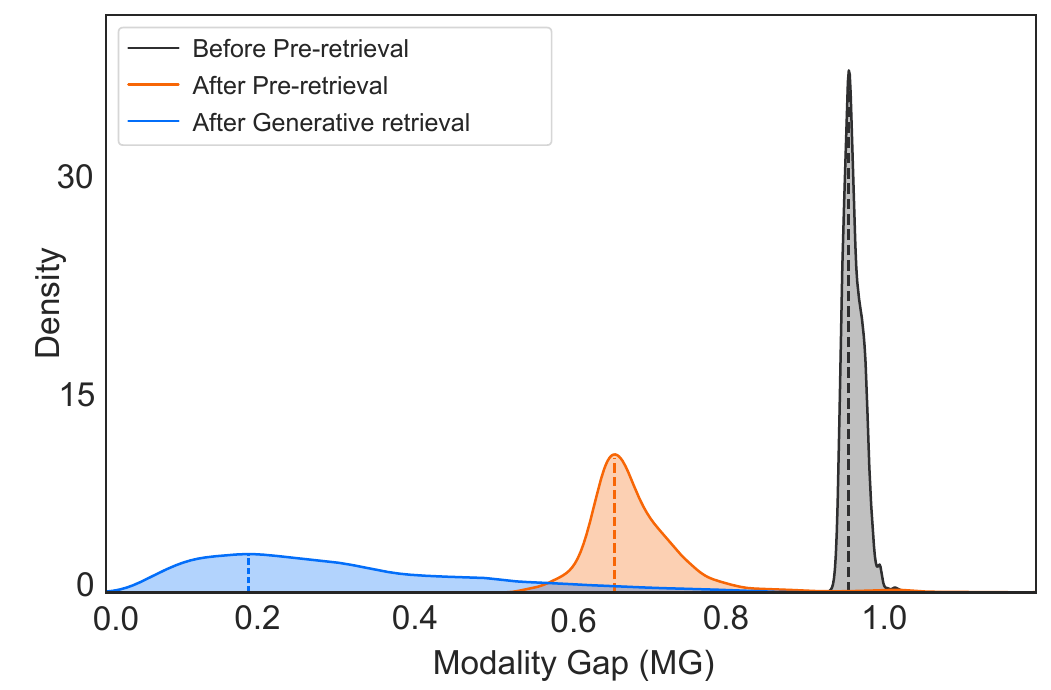}
    \caption{The modality gap distributions on MassSpecGym. A smaller MG indicates better alignment between MS/MS spectra and molecules.}
    \label{fig:Alignment}
    \vspace{-0.5em}
\end{figure}

\subsubsection{Analysis of Modality Alignment}
To evaluate whether our method successfully improves modality alignment between MS/MS spectra and molecules, we define a modality gap metric for each instance $i$ as $\text{MG}(\mathbf{E}_i) = 1 - \cos\left(\mathbf{E}_i^t,\mathbf{E}_i^m\right)$, where $\mathbf{E}_i^m$ is the representation of the ground-truth molecule, and $\mathbf{E}_i^t$ is the representation of the input mass spectrum (before or after pre-retrieval) or the generated molecule (after generative retrieval).
Figure~\ref{fig:Alignment} shows the kernel density estimation~\cite{parzen1962estimation} of the modality gap distribution on MassSpecGym. After the pre-retrieval stage, the distribution shifts leftward compared to the initial distribution, indicating improved alignment through contrastive learning. More notably, in the generative retrieval stage, the modality gap is further and significantly reduced, as the generated molecule is structurally and semantically refined to align closely with the generated molecule.
This progressive narrowing of the modality gap demonstrates that \OURS{} effectively bridges the gap between mass spectra and molecular structures, validating the core advantage of our two-stage framework.

\begin{table}[h]
    
    \centering
    \begin{tabular}{lccc}
    \toprule
    Method
     & MCES $\downarrow$ & \makecell{Morgan \\ Tanimoto} $\uparrow$ & \makecell{RDK \\ Tanimoto} $\uparrow$ \\
    \midrule
    \makecell[l]{SMILES-Trans} &79.39 &0.07 & 0.03 \\
    \makecell[l]{SELFIES-Trans}
 & 33.28 & 0.10 & 0.08\\
     SPEC2MOL &37.76 &0.12 &0.19\\
     MADGEN$_{\text{Pred}}$ & 74.19 &0.08& 0.13\\
    DiffMS & \textbf{18.45} & \textbf{0.28} & \textbf{0.49}\\
    Ours & \underline{21.83} & \underline{0.21} & \underline{0.42}\\
    \bottomrule
    \end{tabular}
    \caption{Molecular generation performance on the test set of MassSpecGym.
    }
    \label{tab:generate}
\end{table}

\begin{table*}[ht]

\centering
\begin{tabular}{lcccccc}
\toprule
\multicolumn{1}{l}{\multirow{2}{*}{\makecell[l]{Library  Type}}} & \multicolumn{1}{c}{\multirow{2}{*}{Method}} & \multicolumn{3}{c}{Recall$\uparrow$ (\%)} & \multicolumn{1}{c}{\multirow{2}{*}{MRR$\uparrow$ (\%)}} & \multicolumn{1}{c}{\multirow{2}{*}{MCES@1$\downarrow$}} \\
\cmidrule(lr){3-5} 
 & & Recall@1 & Recall@5 & Recall@20\\
\midrule
\multirow{3}{*}{\makecell[l]{Weight-based  \\ Retrieval Library}}
 &w/o Generative retrieval & 20.341 & 52.789 & 74.630  & 32.190 & 22.45 \\
 &w/o Pre-retrieval & 41.501 & 59.313  & 73.279  & 49.714 & 18.92 \\
 & \OURS{} (Ours) & \textbf{64.172} & \textbf{72.961} & \textbf{78.782}   & \textbf{67.817} & \textbf{11.14} \\
\midrule
\multirow{3}{*}{\makecell[l]{Formula-based \\ Retrieval Library}}
 &w/o Generative retrieval & 46.030 & 67.925 & 83.202  & 55.900  & 7.83 \\
 &w/o Pre-retrieval  & 52.968 &70.460 & 83.214  & 60.805 & 7.27 \\
 & \OURS{} (Ours) & \textbf{68.478} & \textbf{78.087} & \textbf{84.216} & \textbf{72.472}  & \textbf{5.05} \\
\bottomrule
\end{tabular}
\caption{Ablation study on the two-stage retrieval strategy of \OURS{} on MassSpecGym. The best results are in bold.}
\label{tab:module_ablation}
\end{table*}

\subsubsection{Analysis of Molecular Generation}
While the primary goal of \OURS{} is accurate MS-to-molecule retrieval, the quality of generated molecules, is critical to the final retrieval performance. 
To evaluate the generation capability of our generative model, we employ the MassSpecGym test set and calculate the structural similarity between generated and ground-truth molecules using three metrics: MCES~\cite{curchoe2020all}, Morgan Tanimoto~\cite{vogt2020ccbmlib}, and RDK Tanimoto ~\cite{scalfani2022visualizing}.
We compare our generative model against several methods for de-novo molecule generation from MS/MS spectra, including SMILES Transformer~\cite{sennrich2015neural}, SELFIES Transformer~\cite{krenn2020self}, Spec2Mol~\cite{litsa2021spec2mol}, MADGEN$_{\text{Pred}}$~\cite{wang2025madgen}, and DiffMS~\cite{bohde2025diffms}. 
Their results are taken directly from the original reports of DiffMS and MADGEN, and summarized in Table~\ref{tab:generate}. 
Our generative model achieves competitive performance, ranking second only to DiffMS (SOTA).
We attribute this strong generation quality to two key design choices: (1) a pre-trained spectral encoder derived from the cross-modal contrastive learning, and (2) a context-aware generation framework that conditions the decoder on top-ranked candidate molecules from the pre-retrieval stage.
As a result, the generated molecules are both spectrally consistent and chemically plausible, which in turn enhances the accuracy of the final ranking.

\subsection{Ablation Study}
In this section, we conduct ablation studies to analyze the contribution of key components in \OURS{} and to evaluate the sensitivity of performance to the number of retrieved candidates $K$ in the pre-retrieval stage. Specifically, we first investigate how the pre-retrieval and generative retrieval stages individually contribute to overall retrieval effectiveness. The results are reported in Table~\ref{tab:module_ablation}. 
One can observe that both the pre-retrieval and generative retrieval stages contribute significantly to the performance of \OURS{}.
When used independently, generative retrieval outperforms pre-retrieval alone, indicating that generating a refined molecule is more effective than direct cross-modal matching. However, the best performance is achieved when both stages are combined, demonstrating their complementary nature: the pre-retrieval provides high-quality molecule priors, while the generative retrieval refines these candidates through explicit molecule generation, leading to significantly improved ranking accuracy.

\begin{figure}[ht]
    \centering
    \includegraphics[width=0.5\textwidth]{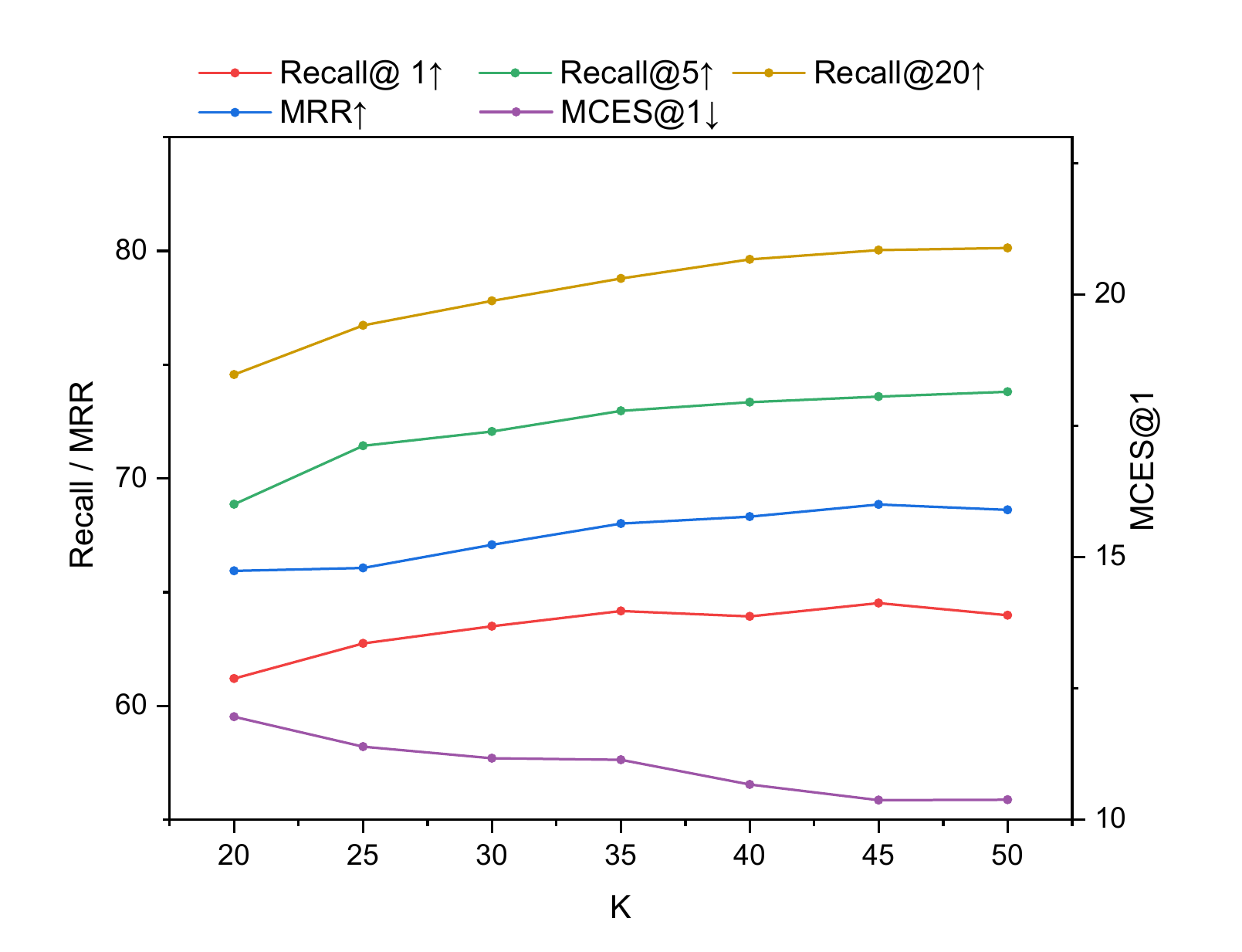}
    \caption{Performance trends with varying $K$. Experiments are conducted on the weight-based retrieval library of MassSpecGym.}
    \label{fig:ab}
    \vspace{-0.5em}
\end{figure}

We further investigate the impact of the number of pre-retrieved molecules ($K$) on retrieval performance. The results (Figure~\ref{fig:ab}) reveal that most metrics reach a plateau when $K>40$. While increasing $K$ beyond this threshold yields marginal improvements in retrieval accuracy, it also introduces higher computational costs. Based on this analysis, we select $K=40$ as the optimal number of pre-retrieved molecules for the pre-retrieval stage, striking a balance between performance gains and computational efficiency.

\section{Conclusion}
In this work, we present \OURS{}, a generative language model-based framework for MS-to-molecule retrieval that addresses the fundamental challenge of cross-modal misalignment between MS/MS spectra and molecular structures. Our two-stage approach, pre-retrieval and generative retrieval, effectively bridges the modality gap by transforming the inherently challenging cross-modal retrieval into a more tractable unimodal molecule retrieval process. Extensive experiments have demonstrated that our method significantly outperforms existing baselines and shows strong generalization. 
Looking forward, we envision several directions for improvement: designing lightweight fusion and generation modules for faster inference; and  incorporating explicit chemical constraints or syntactic rules during generation to enhance molecular structure validity.
By combining generative modeling with retrieval, \OURS{} opens a promising pathway toward accurate, robust, and library-free compound identification in real-world mass spectrometry applications.

\bibliography{aaai2026}

\clearpage
\appendix
\renewcommand{\thefigure}{A\arabic{figure}}
\renewcommand{\thetable}{A\arabic{table}}
\setcounter{figure}{0}
\setcounter{table}{0}

\section*{\huge Appendix}

\section{Construction of \MassRET{}} \label{ap:massret}

Since MassSpecGym~\cite{bushuiev2024massspecgym} already contains data from open-source databases, such as GNPS~\cite{wang2016sharing}, MoNA~\cite{vaniya2019massbank}, and MassBank~\cite{horai2010massbank}, we construct \MassRET{} using the NIST2020~\cite{lemmon2010nist} database, which is a non-open-source database with license restrictions, to verify our model's generalization.

We first standardized the SMILES sequences in these files to ensure a unified and unique representation of compound structures, with those failing the standardization check being removed. Additionally, to avoid data leakage, we removed SMILES that appeared in the MassSpecGym training set.
Next, we further processed the cleaned data by standardizing all mass spectra to relative intensity values. Specifically, for each mass spectrum, each intensity value was divided by the maximum intensity within that spectrum. This normalization process ensures the comparability of mass spectra irrespective of their absolute intensity values, which may vary with different experimental conditions.

The \MassRET{} dataset contains approximately 20k data entries, with a data structure as shown in Table~\ref{tab:MassRET}. Each entry provides spectral information (mzs and intensities), along with the corresponding molecule's SMILES sequence, InChIKey sequence, chemical formula, and molecular weight (parent mass). Additionally, each entry includes metadata related to the mass spectrum, such as the chemical formula and mass-to-charge ratio (m/z) of the precursor ion, adduct type, instrument type used for spectrum generation, and collision energy.

\begin{figure}
    \centering
    \begin{minipage}[c]{0.22\textwidth}
    \includegraphics[width=1\linewidth,trim=4cm 1cm 6cm 2cm,clip]{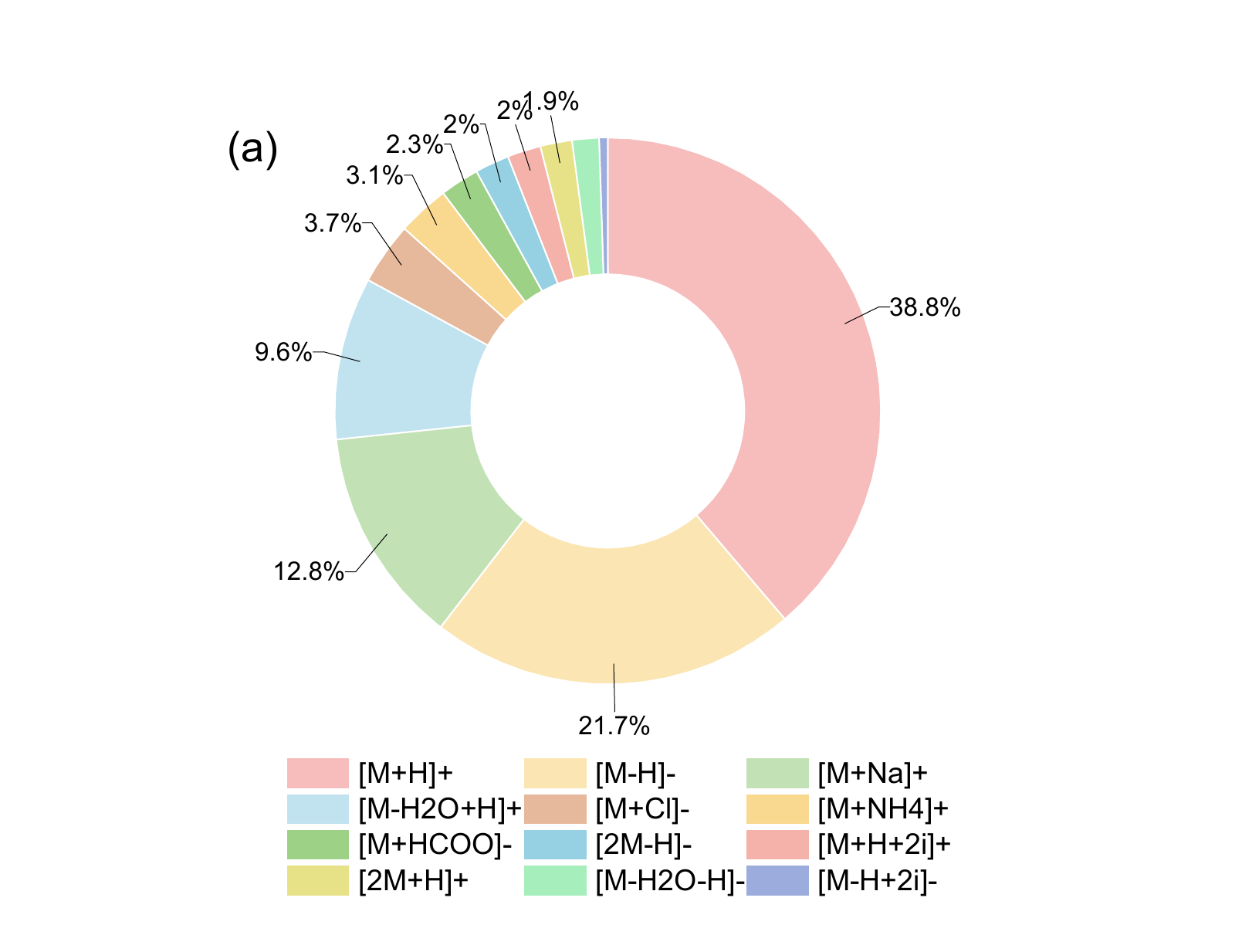}
    \end{minipage}
    \begin{minipage}[c]{0.22\textwidth}
    \includegraphics[width=1\linewidth,trim=4cm 1cm 6cm 2cm,clip]{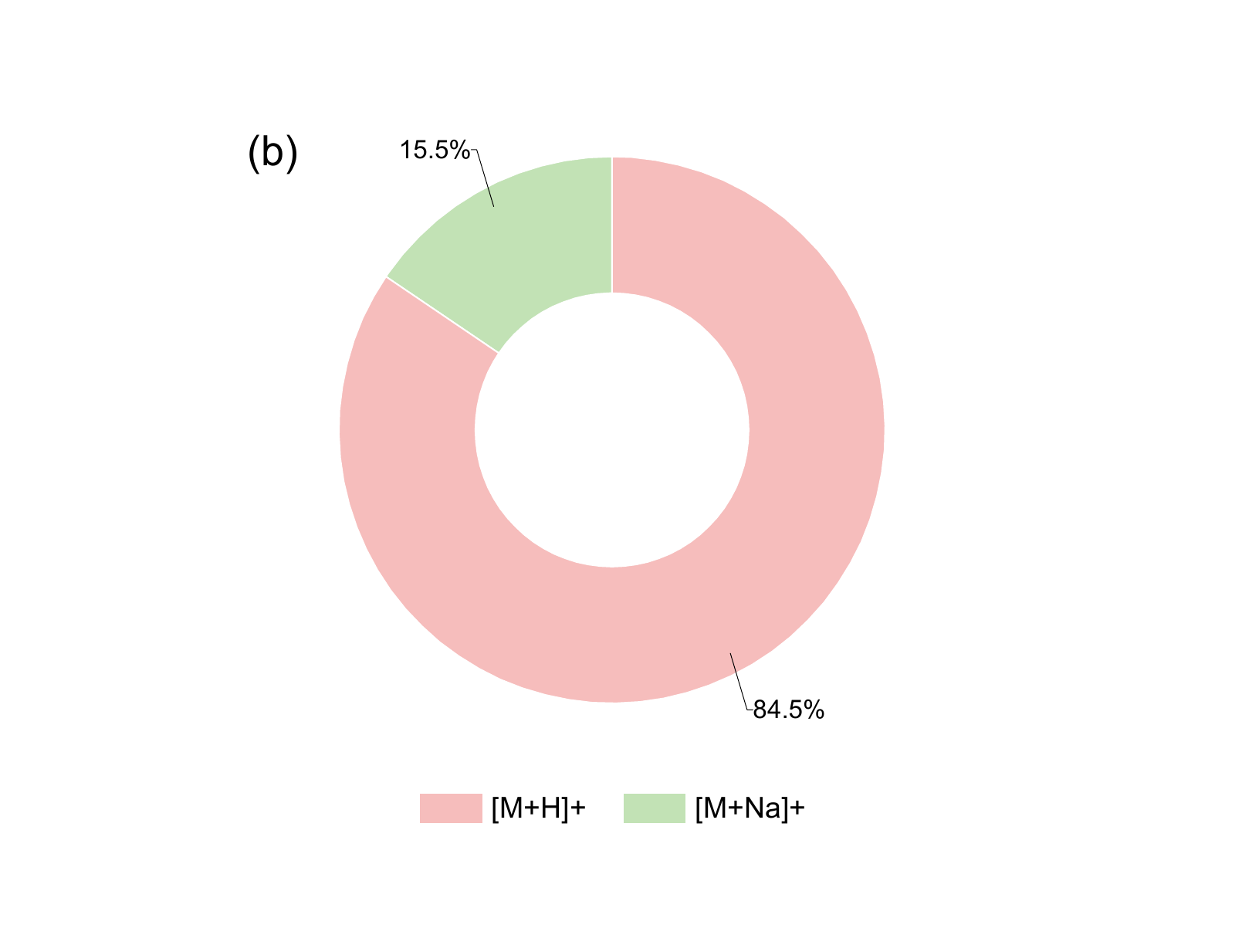}
    \end{minipage}
    \caption{The types and proportions of ionized adducts in (a) \MassRET{} and (b) MassSpecGym.}
    \label{fig:adduct}
\end{figure}

\begin{table}[ht]

\centering
\begin{tabular}{ll}
\toprule
Variable & Description \\
\midrule
identifier & Unique entry identifier\\

\textbf{mzs} & Array of spectrum m/z values\\

\textbf{intensities} & Array of spectrum intensities\\

\textbf{SMILES} & SMILES string of molecule\\

inchikey & 2D InChI key\\

formula & Chemical formula of molecule\\

precursor\_formula & Chemical formula of precursor ion\\

parent\_mass & Mass of molecule\\

precursor\_mz & M/z of precursor ion\\

adduct & Ionization adduct\\

instrument\_type & Type of spectral instrument\\

collision\_energy & Energy of CID fragmentation\\

\bottomrule

\end{tabular}
\caption{Descriptions of all variables present in the \MassRET{} dataset. The key variables are bold.}
\label{tab:MassRET}
\end{table}
In contrast to MassSpecGym, whose metadata only includes two types of ionized adducts and where only 53\% of data entries provide normalized collision energy (leading to data loss), \MassRET{} contains 12 types of ionized adducts and all entries include normalized collision energy, making the included information more complete and better reflective of real-world scenarios. The details are shown in Figure~\ref{fig:adduct}.

\section{Implementation Details of \OURS}
\label{sec:Implementation_Details}
\subsubsection{Model architecture}
The model is composed of a molecular encoder, a spectral encoder, a Spec-Mol Cross-Fusion module, and a molecular decoder. The molecular encoder, derived from the Chemformer~\cite{irwin2022chemformer} encoder architecture, consists of a tokenizer and four bidirectional encoder layers with a feature dimension of 256. It processes SMILES sequences with a maximum length of 512 as input. The spectral encoder, mirroring the molecular encoder's 256-dimensional feature space, is structured with a fully connected layer followed by six multi-head attention encoder layers. It accepts up to 61 tuples, each representing a fragment ion with the first dimension denoting the mass-to-charge ratio (m/z) and the second dimension indicating relative abundance. The cross fusion module employs a cross-attention mechanism~\cite{vaswani2017attention}, using the spectral encoder output as queries and the molecular encoder output as keys and values. The fused representation serves as input to the molecular decoder. The molecular decoder, leveraging the Chemformer decoder architecture, consists of four autoregressive decoder layers and uses beam search~\cite{wiher2022decoding} decoding to incrementally generate SMILES sequences.

\subsubsection{Training detail}
The model is trained in two stages including the pre-retrieval stage and the generative retrieval stage, utilizing spectrum-molecule pairs from the MassSpecGym training dataset. In the pre-retrieval stage, we load the parameters of the molecular encoder and randomly initialize those of the spectral encoder. Contrastive learning is employed for training over 300 epochs, during which the parameters of the molecular encoder are frozen, and only those of the spectral encoder are updated, thus achieving alignment between the molecular and spectral modalities. During contrastive learning, the temperature $\tau$ is set to 0.1 and the number of negative spectral samples $N$ and negative molecular samples $M$ is set to 1. In the generative retrieval stage, we load the pre-trained parameters of the molecular encoder, spectral encoder, and molecular decoder, while randomly initializing the parameters of the cross-fusion module. Here, training proceeds for 30 epochs with the molecular encoder and spectral encoder parameters frozen; only the parameters of the cross fusion module and molecular decoder are updated to enable the decoder to generate the target molecule best matching the current mass spectrum. Both stages utilize the AdamW optimizer and a weight decay of 0.1, setting the learning rate of $1e^{-4}$.

\subsubsection{Inference detail}
The inference process is divided into two stages, with the primary goal of identifying molecules from MassSpecGym's retrieval library that match the input mass spectrum. In the pre-retrieval stage, the inference involves: each molecule in the retrieval library is encoded by the molecular encoder to generate a molecular representation, while the target mass spectrum is encoded by the spectral encoder to produce a spectral representation. The similarity between the mass spectrum representation and each molecular representation is computed, and the $K=40$ most similar molecules are selected as the pre-retrieval results. In the generative retrieval stage, inference includes two steps: molecular generation and re-ranking using the generated molecule. For the former, the size of the beam is set to 5, and the maximum length of the generated sequences is 512. The first molecule is then selected from the generated sequences, encoded by the molecular encoder, and its similarity to the pre-retrieval candidate molecules is calculated. These candidate molecules are then re-ranked based on similarity scores, and top-ranked molecules are selected according to practical requirements.

\section{Introduction of Baselines}
\label{sec:Baselines}
\subsubsection{Fingerprint FFN} Fingerprint FFN~\cite{rumelhart1986parallel} employs a feedforward neural network to predict the target molecule's Morgan fingerprint, followed by sorting candidates based on cosine similarity to the predicted fingerprint. 
\subsubsection{DeepSets} DeepSets~\cite{zaheer2017deep} is evaluated-this model processes mass spectra as sets of raw 2D peak representations. 
\subsubsection{DeepSets + Fourier features} DeepSets + Fourier features~\cite{zaheer2017deep} improves upon DeepSets by enhancing m/z value modeling accuracy through Fourier features. 
\subsubsection{MIST} MIST~\cite{goldman2023mist} first assigns chemical subformulae to spectral peaks via energy-based modeling, then predicts molecular fingerprints using a chemical formula-based Transformer, and ultimately ranks candidates by cosine similarity between fingerprints.
\subsubsection{JESTR} JESTR~\cite{kalia2025jestr} generates molecular representations by integrating molecular graph and molecular fingerprint information, employs spectral binning for spectral representation, and maps both molecular and spectral representations into a shared embedding space. During training, it combines contrastive learning with a candidate molecule regularization strategy. It is the current state-of-the-art method for molecule retrieval from mass spectra.

\section{Introduction of Evaluation Metrics}
\label{sec:Metrics}
\subsubsection{Recall@k}
Recall is a metric that measures the proportion of ground truth molecules retrieved by the model. For a given cut-off point $k$, the Recall@k is defined as:
\begin{align}
\text{Recall@k} = \frac{1}{|M|} \sum_{s=1}^{|M|} \frac{\text{ret}_{m,k}}{\text{rel}_m}
\end{align}
where $|M|$ is the number of queries in the set, $\text{ret}_{m,k}$ is the number of ground truth molecules retrieved for the $m$-th mass spectrum within the top-$k$ results. When the ground truth molecule exists, its value is 1; otherwise, its value is 0. $\text{rel}_m$ is the total number of molecules to be retrieved for the $m$-th mass spectrum.
\subsubsection{Mean Reciprocal Rank(MRR)}
MRR reflects the average rank position of the ground truth
molecule returned in the retrieved molecules. It is computed as follows:
\begin{align}
    \text{MRR} = \frac{1}{|M|} \sum_{m=1}^{|M|} \frac{1}{\text{rank}_m}
\end{align}
where $\text{rank}_m$ is the rank of the ground truth molecule returned for the $m$-th mass spectrum.
\subsubsection{MCES@1}
MCES@1 employs the Maximum Common Edge Subgraph (MCES)~\cite{curchoe2020all} metric to quantify the similarity between the retrieved top molecule $G_1$ and the ground truth molecule $G_t$: 
\begin{align}
    \text{MCES@1} = MCES(G_1,G_t)
\end{align}
Where $MCES$ is the edit distance in the molecular graph. This is an NP-hard~\cite{hochba1997approximation} problem, which we solve via linear programming ~\cite{dantzig2002linear}.

\begin{figure*}[!ht]
    \centering
    \includegraphics[width=1\linewidth,trim=3cm 2cm 3cm 2cm,clip]{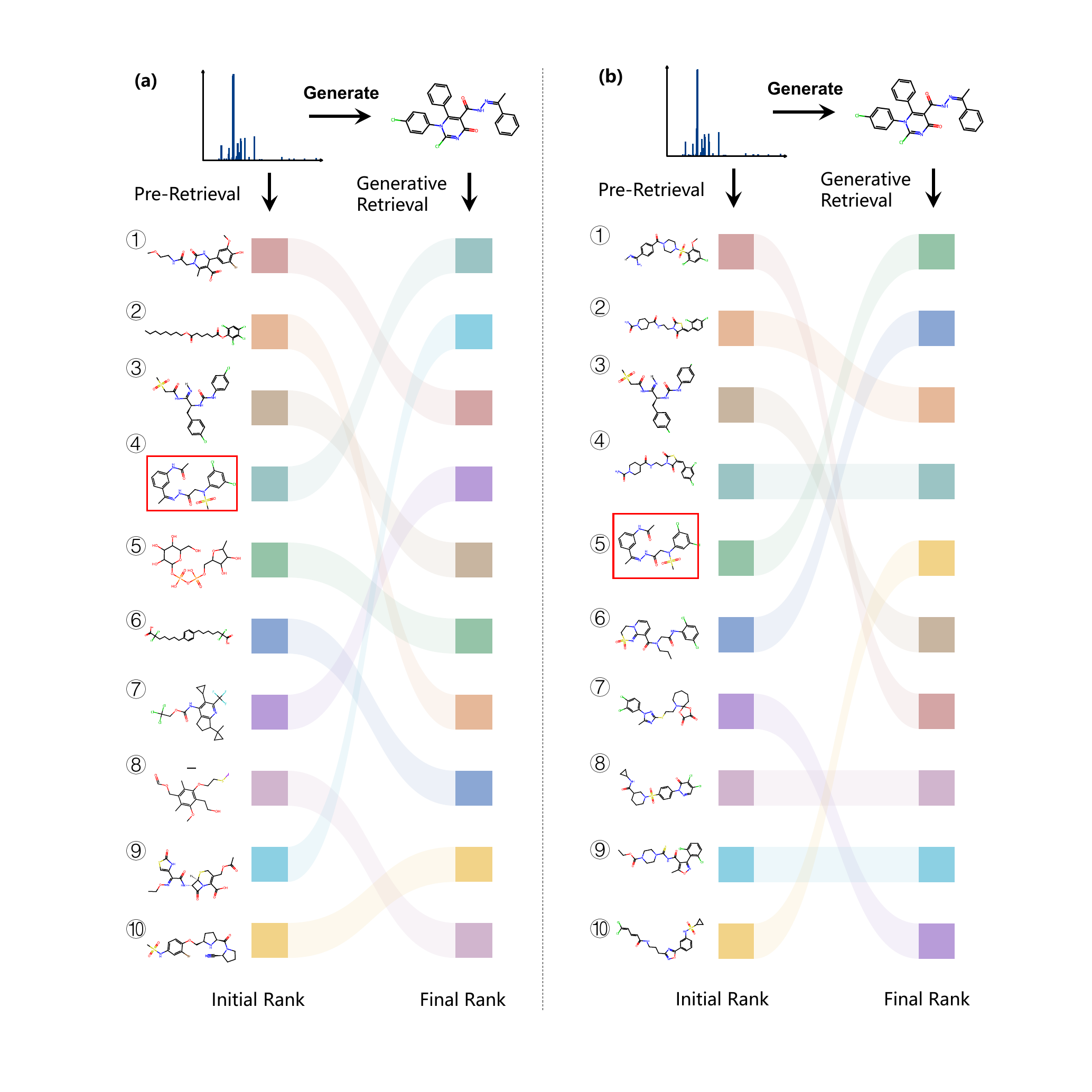}
    \caption{Visualization of top-10 retrieval results of \OURS{} including the pre-retrieval stage and the generative retrieval stage. The target molecule (a) moves from the 4th to the 1st after re-ranking on the weight-based retrieval library and (b) climbs from 5th to 1st rank after re-ranking on the formula-based retrieval library. } 
    \label{fig:case1}
\end{figure*}

\begin{table*}[!ht]
\centering

\begin{tabular}{lcccccc}
\toprule
\multicolumn{1}{l}{\multirow{2}{*}{\makecell[l]{Library  Type}}} & \multicolumn{1}{c}{\multirow{2}{*}{Method}} & \multicolumn{3}{c}{Recall$\uparrow$} & \multicolumn{1}{c}{\multirow{2}{*}{MRR$\uparrow$}} & \multicolumn{1}{c}{\multirow{2}{*}{MCES @ 1$\downarrow$}} \\
\cmidrule(lr){3-5} 
& & Recall@1 & Recall@5 & Recall@20\\
\midrule

\multirow{3}{*}{\makecell[l]{Weight-based  \\ Retrieval Library}} 
&w/o Generative retrieval & 15.234 & 43.438  & 69.871 & 27.921 & 26.90  \\
&w/o Pre-retrieval & 34.563 & 51.652 & 69.293 & 42.788 & 19.21  \\
& \OURS{} (Ours) & \textbf{54.042} & \textbf{64.347} & \textbf{72.984}  &\textbf{58.835} & \textbf{12.08}   \\
\midrule
\multirow{3}{*}{\makecell[l]{Formula-based \\ Retrieval Library}} 
&w/o Generative retrieval & 36.448 & 55.199 & 71.275 & 44.896 & 9.24 \\
&w/o Pre-retrieval & 42.065 & 55.891 & 70.501 & 48.711 & 8.48 \\
& \OURS{} (Ours) & \textbf{51.141} & \textbf{60.062} & \textbf{70.671}  & \textbf{55.565}  & \textbf{6.94} \\
\bottomrule
\end{tabular}
\caption{Ablation study on the two-stage retrieval strategy of \OURS{} on \textbf{\MassRET{}}. The best results are in bold.}
\label{tab:ab2}
\end{table*}

\section{Visualization of Retrieval Process}
To further understand \OURS{}, we visualize its workflow, which involves using the input mass spectrum to perform initial ranking in the pre-retrieval stage, and generating a refined molecule in the generative retrieval stage to conduct re-ranking. For better illustration, in the pre-retrieval stage, we select the top-10 retrieved molecules as contextual priors to guide the generation of refined molecules in the generative retrieval stage.

We take \texttt{ID0206445} from the MassSpecGym dataset as an example (i.e., the input mass spectrum). Figure ~\ref{fig:case1} shows the retrieval results on the weight-based retrieval library and the formula-based retrieval library. This case highlights how the generative refinement process effectively corrects alignment errors and enhances retrieval accuracy.

\section{Additional Ablation Experiments}
Besides MassSpecGym, we perform ablation studies on \MassRET{} to further verify the contributions of the pre-retrieval and generative retrieval stages. The results are shown in Table~\ref{tab:ab2}.
Our findings align with those from MassSpecGym: both the pre-retrieval and generative retrieval stages contribute significantly to the performance of \OURS{}.

\section{Efficiency Comparison}
\begin{table}[!ht]
    
    \centering
    \begin{tabular}{lcc}
        \toprule
        Method &  Time (s) & Param. (M) \\
        \midrule
        DeepSets  & 0.33 &2.6  \\
        Fingerprint FFN  & 0.36 &2.6  \\
        DeepSets + Fourier & 0.40 &7.8  \\
        MIST   & 0.64 &98.6\\
        JESTR   & 0.57 &19.9\\
        \OURS{} (Ours)   & 0.41+0.69=1.10 &13.6\\
        \bottomrule
    \end{tabular}
    \caption{Comparison of the average retrieval time per instance on MassSpecGym.}
    \label{tab:efficiency}
\end{table}

Table~\ref{tab:efficiency} presents the average retrieval time per instance on MassSpecGym and the number of model parameters. The pre-retrieval stage of \OURS{} requires 0.41s, which is longer than that of DeepSets and Fingerprint FFN but shorter than that of DeepSets + Fourier, MIST, and JESTR. The retrieval approach in the pre-retrieval stage is consistent with that of other models, which performs retrieval by computing the similarity between molecular embeddings and spectral embeddings. In this scenario, the retrieval time is positively correlated with the model's parameter count. Owing to their extremely small parameter counts, DeepSets and Fingerprint FFN exhibit very short retrieval times but poor retrieval performance. The pre-retrieval stage has fewer parameters than MIST and JESTR (resulting in less retrieval time than they do); however, its retrieval performance is on par with theirs. The generative retrieval stage of \OURS{} takes 0.69s, which involves generating a refined molecule from the MS/MS spectrum and subsequently calculating the similarity between the two molecular embeddings for retrieval. Compared with the pre-retrieval stage, this stage includes an additional step of molecule generation, rendering it more time-consuming.

\end{document}